\documentclass[a4paper,fleqn]{cas-dc}

\usepackage[numbers]{natbib}
\usepackage{hhline} 
\usepackage{graphicx}
\usepackage{adjustbox}
\usepackage{amsmath,amssymb,amsfonts}
\usepackage{mathtools}
\usepackage{boldline} 
\usepackage{titlesec}
\usepackage{titlesec}
\titleformat*{\section}{\normalsize\bfseries}
\usepackage[utf8]{inputenc}
\usepackage{url}

\ExplSyntaxOn \cs_gset:Npn \__first_footerline: { \group_begin: \group_end: } \ExplSyntaxOff

\titlespacing*{\section}
{0pt}{3ex plus .2ex}{2ex plus .2ex}

\titlespacing*{\subsection}
{0pt}{3ex plus .2ex}{2ex plus .2ex}

\titlespacing*{\subsubsection}
{0pt}{3ex plus .2ex}{2ex plus .2ex}

\begin{document}
\let\WriteBookmarks\relax
\def\floatpagepagefraction{1}
\def\textpagefraction{.001}

\shorttitle{}


\title [mode = title]{Are Deep Learning Classification Results Obtained on CT Scans Fair and Interpretable?}                      



%
\author[1]{Mohamad M.A. Ashames}


\author[1]{Ahmet Demir}
\author[2]{Omer N. Gerek}
\author[4]{Mehmet Fidan}
\author[1]{M. Bilginer Gulmezoglu}
\author[1]{Semih Ergin}
\author[3]{Mehmet Koc}
\author[2]{Atalay Barkana}

\author[5]{Cuneyt Calisir}

\affiliation[1] {addressline={Department of Electrical and Electronics Engineering}, 
    organization={Eskisehir Osmangazi University},
    city={Eskisehir},
    postcode={26480}, 
    country={Turkey}}

\affiliation[2] {addressline={Department of Electrical and Electronics Engineering}, 
    organization={Eskisehir Technical University},
    city={Eskisehir},
    postcode={26555},
    country={Turkey}}

\affiliation[3] {addressline={Department of Computer Engineering}, 
    organization={Eskisehir Technical University},
    city={Eskisehir},
    postcode={26555},
    country={Turkey}}

\affiliation[4] {addressline={Vocational School of Transportation}, 
    organization={Eskisehir Technical University},
    city={Eskisehir},
    postcode={26555},
    country={Turkey}}

\affiliation[5] {addressline={Department of Radiology}, 
    organization={Manisa Celal Bayar University},
    city={Manisa},
    postcode={26480},
    country={Turkey}}




\begin{abstract}
Following the great success of various deep learning methods in image and object classification, the biomedical image processing society is also overwhelmed with their applications to various automatic diagnosis cases. Unfortunately, most of the deep learning-based classification attempts in the literature solely focus on the aim of extreme accuracy scores, without considering interpretability, or patient-wise separation of training and test data. 
For example, most lung nodule classification papers using deep learning randomly shuffle data and split it into training, validation, and test sets, causing certain images from the CT scan of a person to be in the training set, while other images of the exact same person to be in the validation or testing image sets. This can result in reporting misleading accuracy rates and the learning of irrelevant features, ultimately reducing the real-life usability of these models.
When the deep neural networks trained on the traditional, unfair data shuffling method are challenged with new patient images, it is observed that the trained models perform poorly. In contrast, deep neural networks trained with strict patient-level separation maintain their accuracy rates even when new patient images are tested. Heat-map visualizations of the activations of the deep neural networks trained with strict patient-level separation indicate a higher degree of focus on the relevant nodules. We argue that the research question posed in the title has a positive answer only if the deep neural networks are trained with images of patients that are strictly isolated from the validation and testing patient sets. 

\end{abstract}




\setcitestyle{square}

\maketitle

\section{Introduction}

The society of biomedical image processing has an abundance of image and object classification publications due to the great success of various deep learning methods. The biomedical images in various automatic diagnostic cases may consist of stand-alone image outputs such as X-rays. However, a majority of handled image data contains outputs in terms of batch scans, CT and MRI are typical examples. In a single batch, scans from slightly different offsets are obtained in order to observe the same part of the same person. Deep learning has been shown as a prevalent and effective algorithm in the diagnosis of many medical images \cite{Song2017}. On the other hand, it has also been criticized that deep learning is not reliable because it is not truly explicable~\cite{Pretz2022}. The method may work, yet it may be impossible to understand totally the underlying reason. Consequently, its continuing accuracy for any new diagnosis case is never guaranteed. Furthermore, while the method is slowly learning in the training phase from a new database, a random portion of its learned memories may abruptly fail~\cite{Pretz2022}. 

Separating data into train-test sets is required to determine the performance of a machine learning (ML) algorithm in the case of supervised and semi-supervised learning. For this purpose, a dataset is taken and divided into two subsets, preferably in a random way. One of these subsets is utilized to adapt the algorithm parameters, and that set is defined as the training set. The features of the exclusive subset (i.e., the set which is not used in the training process) are applied to the algorithm as an input to make a valid success assessment. This excluded subset is defined as the test set. If the sample data is sufficiently large, one can split the whole data into training and test sets and still obtain a large number of samples, both for training and testing. If the data count is relatively small, the remedies include methods such as a modern approach of reinforcement learning, or a more classical approach of cross-folding train/test data.

For example, Goodfellow et al. asserted that the training and testing data could be generated with a probability distribution over datasets which is called the data generation process~\cite{Goodfellow2016}. As a rule, the independent and identically distributed (i.i.d.) assumption of the data is critical, meaning that the samples in each data set are independent of each other, and the training and test sets are equally distributed. 

Unfortunately, many of the ML-based diagnosis attempts in the literature did not handle image datasets that are obtained from batch scans with sufficient care regarding the {\em independence} condition as explained above. In a majority of the cases, the test-train separation of images from multiple scans was done randomly, providing images from exactly the same scan to appear in the training as well as the test or validation sets. Since such a situation is a direct violation of the independence requirements, we investigate the effect of such {\em unfair} train-test splitting on the performance of ML methods in terms of detection accuracy and overall algorithmic interpretability. Besides, the efficiency and interpretability improvement under the strict (i.e., patient-wise) separation of train and test (or validation) data splitting case is studied in this work.

As a test case of the careful test-train separation problem, we consider malignancy detection of lung nodules from computed tomography (CT) scans, where the literature is crowded with several deep learning algorithm results. 

In a survey paper ~\cite{Gu2021}, Gu et al. review available CAD systems applying deep learning to CT scan data for lung nodule detection, segmentation, classification, and retrieval. They argue the advance of deep learning, define various important characteristics of lung nodule CAD systems, and evaluate the performance of certain studies against different databases such as LIDC, LIDC-IDRI, LUNA16, DSB2017, NLST, TianChi, and ELCAP. In the selected classification studies, the accuracy rates range from 75.01\% to 98.31\%. High accuracy results arise from the inclusion of different CT images belonging to the same patient in both training and test sets. Throughout the paper, we call this the UNFAIR case. In this case, only image-wise cross-fold validation technique is used \cite{Song2017,Wu2018,Xie2017a,Xie2017b,Abbas2017,Zhao2018,Xie2018,Shen2017,Sahu2018,Hussein2017,Potghan2018,Liu2017,Nibali2017}. In these studies, LIDC-IDRI database is widely used with various classification methods such as convolutional neural network (CNN)~\cite{Song2017}, an interpretable and multi-task learning CNN ~\cite{Wu2018}, three pre-trained ResNet-50 models~\cite{Xie2017a},  multi-view knowledge-based collaborative (MV-KBC) deep model~\cite{Xie2017b},  CNN, RNN and softmax~\cite{Abbas2017}, forward and backward GAN (F\&BGAN) and Multi-scale VGG16 (M-VGG16) network~\cite{Zhao2018}, algorithm that fuses the texture, shape and deep model-learned information (Fuse-TSD)~\cite{Xie2018}, multi-crop Convolutional Neural Network (MC-CNN)~\cite{Shen2017}, lightweight and multiple view sampling based Multi-section CNN architecture~\cite{Sahu2018}, end-to-end deep multi-view CNN~\cite{Hussein2017}, K-Nearest Neighbor (kNN) and Multi-Layer Perceptron (MLP)~\cite{Potghan2018}, multi-view CNN (MV-CNN)~\cite{Liu2017}, and obtained accuracy rates vary between 84.15\% and 98.31\%. Nibali et al. assessed the usefulness of very deep convolutional neural networks in the expert-level classification of malignant lung nodules~\cite{Nibali2017}. Based on the well-known ResNet architecture, they investigated the effect of curriculum learning, transfer learning, and different network depths on malignancy classification accuracy; and obtained an accuracy rate of 89.9\% for the LIDC-IDRI database. In~\cite{Shaffie2019}, only the LIDC database was used with denoising autoencoder (DAE) and 3D Resolved Ambiguity Local Binary Pattern (3D-RALBP) methods, and a maximum accuracy rate of 94.95\% was obtained. The Optimal Deep Neural Network (ODNN) was applied to CT images and then, optimized with the Modified Gravitational Search Algorithm (MGSA) to determine the classification of lung cancer, and the accuracy of 94.56\% was obtained for ELCAP database~\cite{Laksh2019}. When the CT images taken from the Cancer Imaging Archive were used for lung nodule classification, extreme accuracy rates of 99.51\% and 97.14\% were obtained using kNN with AlexNet \& mRMR feature extractor in \cite{Togacar2020}, and LDA classifier in \cite{Aggarwal2015} respectively. In another study, Tran et al. suggested a new 2D architecture for a deep convolutional neural network using focal loss, and they obtained a high accuracy rate of 97.2\% for the LUNA16 database~\cite{Tran2019}.

The extreme accuracy rates mentioned above {\em could} be attributed to a possible overfit due to an unintentional leak of same-batch image data to both training and test sets. All of these manuscripts report that the whole image lot was shuffled completely, and train-test separation was done randomly. In such a case, it is perfectly possible that some images from the same patient scan may go to the training set, while the rest may go to the test set, making an unfair splitting that is prone to overfitting with too high accuracy results.

Contrary to the above-mentioned unfair splitting, a fair splitting approach is also possible, where one must carefully assign distinct patients' scan images to train and test sets. The literature also contains several papers, where this attention was paid~\cite{Kumar2015,Liao2019,Gruet2019,Paul2018,Utkin2019}. In these studies, the maximum accuracy rates of 75.01\%, 81.47\%, 82.1\%, 89.45\%, and 91.8\% were obtained by using deep features extracted from an autoencoder along with a binary decision tree \cite{Kumar2015} for a part of LIDC database, a 3-D version of the RPN using a modified U-net and a leaky noisy-OR model for DSB2017 database~\cite{Liao2019}, CNN model for LIDC-IDRI database \cite{Gruet2019}, VGG-s CNN models for NLST database \cite{Paul2018} and an ensemble of triplet neural networks for LUNA 16 database \cite{Utkin2019} respectively. In \cite{Coruh2021} and \cite{Teramoto2017}, authors created their own databases, and the accuracy rates of 75.2\% and 71.1\% were obtained using artificial intelligence (AI) systems obtained from the union of convolutional neural networks (CNN) and 2D deep convolutional neural network architecture respectively.

Although overfitting due to unfair test-train dataset splitting seemingly gives higher accuracy results, the reliability of the results could be questionable from the following aspects:
\begin{itemize}
    \item Do these trained ML techniques still provide high accuracy for a completely new “challenge" data set?
    \item Do these trained ML techniques perform classifications by really focusing on the actual nodule positions (marked by radiologist experts)?
    \item Hence, are these techniques {\em interpretable}?
\end{itemize}
A follow-up question automatically arises:
\begin{itemize}
    \item If we perform strictly fair test-train splitting, does this improve performance on the challenge data set and interpretability?
\end{itemize}

This study provides experiments comparing the reliability of deep learning algorithms for lung nodule classification by implementing fair and unfair data splitting. Since the datasets from the LIDC-IDRI database have been widely used for studying nodule detection and classification methods, including various studies relevant to this work, the LIDC-IDRI database is used in the experimental studies of this paper. 

The comprehensive review by Loizidou et al. in a different case of detection and classification in mammography clearly points out the problems that arise when strict patient-wise training/validation/test separation is not performed~\cite{Loizidou}. They propose that images and image labels (i.e., ROIs) of the same patient should be incorporated into the training, validation, or test mammography datasets. They also express concern regarding the high classification accuracy rates reported in various papers that failed to perform this separation, as they render the performances unverifiable for new patient cases in real-life. In this study, we explore this idea in the context of CT scans, demonstrating the invalidity of unfair training accuracy results numerically. Furthermore, we show that deep neural networks trained using unfair random image splitting are incapable of focusing attention on indicator regions of CT images (i.e., nodule regions), which renders the results completely non-interpretable. Several experimental studies related to unfair and fair data splitting cases for lung nodule classification are performed. For this purpose, deep learning neural network with three architectures (mobileNetV2, efficientNet, and VGG16). For all of these deep learning methods, the model evaluation with a new patient dataset demonstrates that data shuffling done inattentively makes the trained model inapplicable in real-life, as well as reducing the learning capability of the model by focusing on irrelevant features in the neural network layers~\cite{Jekova2021}. On the contrary, strict patient isolation between train and test datasets provides significantly better results in real-life challenge datasets containing images from new patients. Besides, this isolation helps deep neural network layers to better focus their attention on the correct nodule locations on the image. This interpretability attribute is visualized with a heat-map technique, which renders higher activation network portions red while rendering the low activation portions blue. Finally, this visualization is further quantized as a numerical value to make an assessment of interpretability using three novel interpretability functions introduced herein.

\section{Materials and Method}
\subsection{Dataset}

The dataset used in the experimental part of our proposed approach is extracted from the publicly available LIDC/IDRI dataset~\cite{Armato2011}. National Cancer Institute (NCI) started to create the LIDC database in 2001, and the Foundation for the National Institutes of Health (FNIH) supported it to create a bigger database named LIDC/IDRI in 2004. LIDC was supported by five academic medical centers, and two more centers came with the addition of IDRI.

LIDC/IDRI is one of the largest available databases as it contains 1018 thoracic CT scans taken from 1010 different patients.  These scans are acquired by using a number of different scanner devices and acquisition parameters. Each scan in the dataset has an XML (eXtensible Markup Language) file that contains diagnosis and nodule reports created by four experienced radiologists. These reports are created in two phases: blinded and unblinded reading phases. In the blinded reading phase, radiologists independently classify each nodule into three categories (3mm~$\leq$ nodule $ < $~30mm, nodule~$<$~3mm, non-nodule $>$ 3mm) according to the nodule diameters. In the unblinded reading phase, each radiologist sees the blinded phase decisions of the other three radiologists anonymously and specifies his/her final decision about a nodule. Radiologists are not expected to achieve any consensus in this process. Since the probability of being in the malignant class for the nodules having a diameter greater than or equal to 3 mm is higher compared to other nodules, the main goal of the LIDC/IDRI project is to determine the nodules which are in this category. Therefore, radiologists are asked to draw the nodule contours, specify their locations and give malignancy scores only for these nodules. All these data are saved to the XML files of scans and used as ground truths in further studies.

\begin{figure} [!h]
	\centering
	\includegraphics[scale=.75]{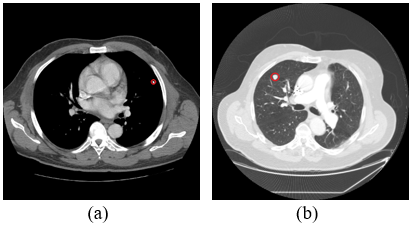}
	\caption{Axial CT images of (a) Benign, (b) Malignant pulmonary nodules. Nodules are circled with red color on the images.}
	\label{figsamples1}    
\end{figure}

In this study, we utilized the LIDC/IDRI dataset to investigate the malignancy of pulmonary nodules in Computed Tomography (CT) scans. The selection of the CT scans was based on the “LIDC Nodule Size List” document, which was available on the official website of the Cancer Imaging Archive. The document provided information on the number of nodules, the number of radiologists who identified each nodule, and the nodule volumes for each scan. The malignancy scores for each nodule were determined by collecting data from the XML files accompanying each scan in the dataset folder. The nodule ID information was used to identify the malignancy characteristics of a nodule in a scan, as each radiologist gave a different name to the same nodule. To ensure the reliability of the dataset, only nodules that were scored by at least three radiologists were used in the study, and each selected nodule was re-examined and approved by a practicing radiologist. The final malignancy score was determined by averaging the scores assigned by each radiologist. Nodules with an average score of less than or equal to 1.5 were classified as benign, and those with an average score of larger than or equal to 3.5 were classified as malignant. A total of 63 benign nodules and 98 malignant nodules were included in the study. The nodule images were acquired by using the noduleID values of the selected nodules and the “imageZposition” parameter, which determined the slice numbers of each nodule in a scan. The images were in $512\times 512$ DICOM format, and the MicroDicom software was used to display, analyze, and convert the selected images into $512\times 512$ PNG format. A total of 303 benign and 919 malignant class images were acquired, and Figure~\ref{figsamples1} presents a sample of a benign and a malignant lung CT image.

To enhance the generalizability of the trained networks and avoid overfitting, we performed data augmentation on the selected original dataset. Each image in the benign and malignant classes was rotated by +2, -2, +4, and -4 degrees, resulting in a total of 1515 benign and 4595 malignant images. After augmenting the dataset, we split the data into two categories: unfair data splitting and fair data splitting. The former refers to a dataset where images from the same scan can be used in both the training and testing processes. We randomly divided the augmented dataset into train, validation, and test sets, without applying patient-wise division. The train folder comprised 969 benign and 2940 malignant images, the validation folder contained 410 benign and 1241 malignant images, and the test folder had 136 benign and 414 malignant images. In contrast, the fair dataset was created by implementing patient-wise data splitting. We separated the data into a train-validation folder and a test folder. CT scans belonging to patients in the train-validation folder were solely used for training and validation to prevent any correlation between the images in the train-validation and test folders. Hence, the test dataset did not contain any images that were used to train and validate the models.

\subsection{Deep Neural Network Architectures}

 Nowadays, deep neural networks (DNNs) have become the gold standard in classification problems, and huge portions of these networks are composed of CNNs. In this paper, the classification task is realized by using three well-known DNN architectures, and short explanations of these architectures are given below.

Simonyan and Zisserman studying at the Visual Geometry Group Lab of Oxford University have suggested VGG-16 architecture in 2014~\cite{Simonyan2015}. VGG-16 Network architecture contains 16 groups of layers in total. It takes RGB images with a resolution of 224×224 pixels as input. It has a convolution kernel with the size of 3×3 and a maximum pooling layer with the size of 2×2. It is one of the most widely used architectures in  various pattern recognition studies in spite of its comparatively slower training process. 

EfficientNet is another DNN architecture that scales some parameters such as depth, width, and resolution with the help of a compound coefficient~\cite{Tan2019}. EfficientNet differs from the other architectures by uniformly balancing these parameters. It aims to lower the calculation cost by dividing the conventional convolution into two phases. Along with that, it diminishes possible losses resulting from the usage of Rectified Linear Unit (ReLU) by utilizing a linear activation function at its final layer blocks.

MobileNet is a newly invented neural network architecture by a number of Google researchers, and it is adapted mainly to mobile devices~\cite{Sandler2018}. Since many mobile devices
have some source limitations, researchers find them attractive due to their fruitful characteristics, such as being small and low-latent. MobileNetV2 is the second version of MobileNet, and some bottleneck layers are used. Also, MobileNetV2 does a filtering operation on the features to overcome the nonlinearity problem.

\subsection{Experimental Study}

Google Colaboratory or “Colab” was used as an environment for implementing our experiments. The environment provides a tool for writing and executing python code and is especially applicable for machine learning tasks \cite{colab}.
Keras with Tensorflow backend is used to import the DNN architectures. End-to-end binary classification is carried out by modifying all three ImageNet pre-trained DNN final layers with binary softmax layers and by training them. A simple resizing operation on the dataset images is carried out according to the default input size of the networks before giving them as input to the DNNs. 

\subsubsection{Training Procedure}

Training parameters used in the experiments are given in Table 1. The table shows the starting value of the learning rate, and it is reduced by one-tenth if no validation accuracy improvement is seen for a number of epochs.

 \begin{table}[!b]
\centering
\caption{DNN training parameter settings}
\label{tab:table1}
\begin{tabular}{ c | c }

\hlineB{2}
\textbf{Parameter} & \textbf{Value} \\ \hlineB{2}
Learning Rate & 0.0001 \\
Epoch Number & 50-200 \\
Batch Size & 32 \\
Optimizer & ADAM \\ \hlineB{2}
\end{tabular}
\end{table}

\begin{figure*} [t!]
	\centering
	\includegraphics[width=.9\textwidth]{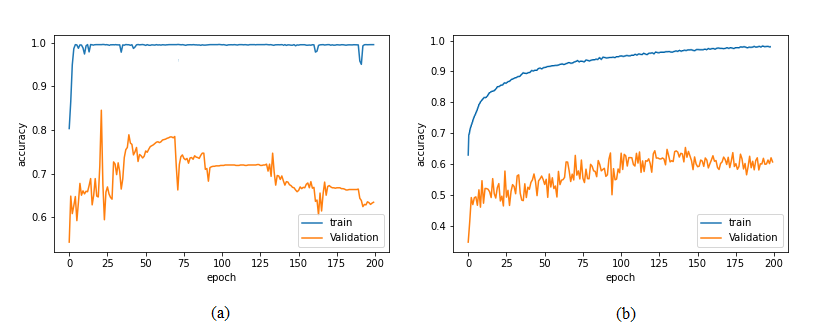}
	\caption{Training and validation accuracies for both (a) unfair and (b) fair train-validation procedures using EfficientNet.}
	\label{figsamplesacc}    
\end{figure*}

In the unfair train-validation process, 70\% of the dataset is utilized for training and validation, while the remaining 30\% is set aside for testing. This method involves feeding different images from the same CT scan into the input layers of the architectures, resulting in unfair training. Similarly, the images in the test set could also be from the same patients utilized in the train-validation process, resulting in unfair testing. Such train-validation-test sets contain images from all CT scans, producing misleading accuracy values and causing the models to overfit at the early training stages, as demonstrated in Figure~\ref{figsamplesacc}-a.

\begin{figure} [h]
	\centering
	\includegraphics[scale=.21]{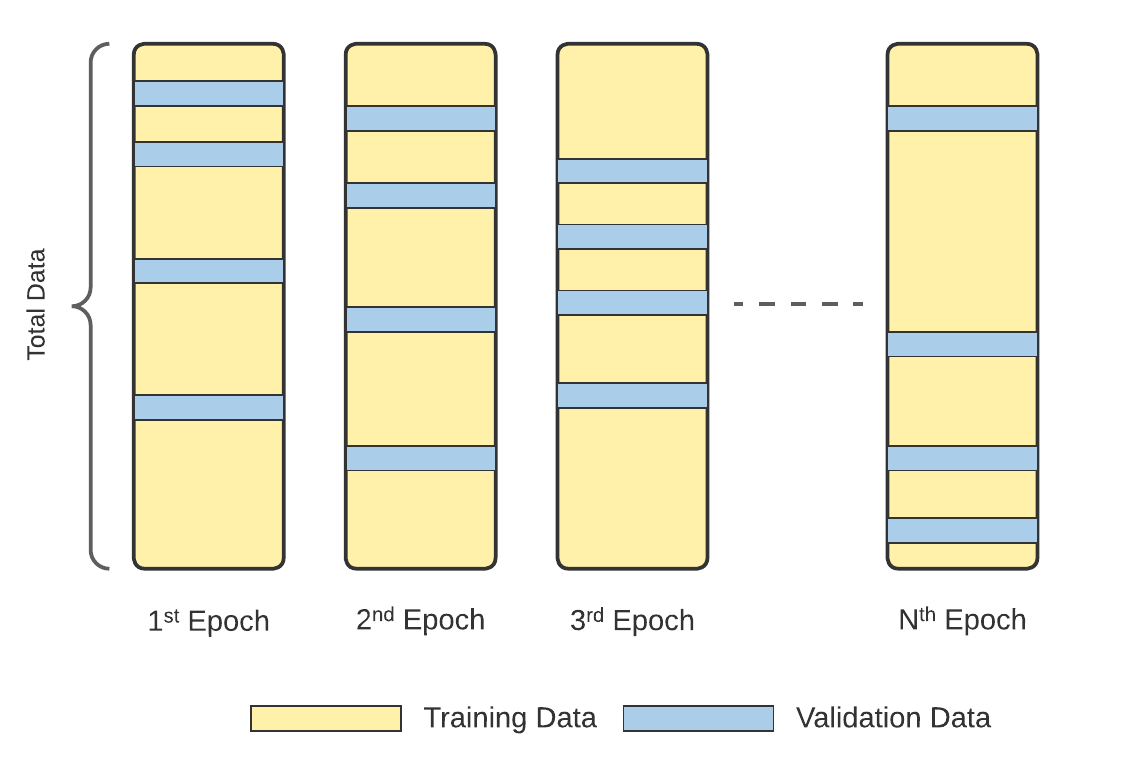}
	\caption{Monte Carlo Cross-Validation. The diagram shows that patients of the validation set change each time a new epoch starts.}
	\label{figsamples2}    
\end{figure}

 In order to avoid overfitting and report reliable accuracy results, CT scans of different patients were divided into separate folders for the second experimental set, which we call the FAIR training procedure. Monte Carlo Cross Validation (MCCV)~\cite{Xu2001} was applied to use CT scans belonging to different patients for each training and validation set in the architectures. Images from a random group of patients are used for training, while images from the rest of the patients are used for validation. Furthermore, images from a completely different set of patients are used for testing. The patient-wise train-validation splitting in MCCV is illustrated in Figure~\ref{figsamples2}. The improvement in the learning process and validity of the reported accuracy results are analyzed. Figure~\ref{figsamplesacc}-b clearly shows that the proposed training process improves in time and no inconsistent overfitting occurs. Furthermore, the resultant networks provide accuracy results that are more reliable, as will be discussed in Sec. 2.3.2.

\subsubsection{Classification Results}

 Three DNN architectures; MobileNetV2, EfficientNetB0, and VGG16, were trained and validated, first through the unfair training-validation separation, and then through fair dataset splitting by MCCV. Table~\ref{tab:table2} compares the classification accuracies for the unfair and fair experiments of each architecture. As expected, the architectures tend to report misleadingly high accuracies when they're unfairly trained and tested, while they reach lower (but actually correct) accuracy values when patient-wise data splitting is carried out, and different CT scans are used for testing.

In order to assess the correctness and validity of the reported test accuracies, CT images of a completely isolated set of patients (called the challenge set) were applied to the trained networks. The obvious observation is that the reported test accuracies (left-side column) of the unfairly trained network are far from being valid for the challenge set (right-side column), whereas the performance of the fair-trained network is totally consistent with the reported test accuracies. Interestingly, certain networks (i.e., EfficientNet and VGG16) result in an extreme failure in the challenge dataset when they are unfairly trained, giving an impression that overfitting and patient-learning could be a more pronounced issue in these networks. In order to avoid that situation, when the networks are fairly trained, the test and challenge accuracies could become modestly high, consistent, and reliable.

\begin{table*}[!h]
\centering
\caption{Classification accuracies obtained by implementing fair and unfair training-testing for both test and challenge datasets.}
\label{tab:table2}
\begin{tabular}{l|c|cc|cc}
\hlineB{2}
\multirow{2}{*}{DNN Architecture} & \multicolumn{1}{l|}{\multirow{2}{*}{Number of Epochs}} & \multicolumn{2}{c|}{Fair}                                           & \multicolumn{2}{c}{Unfair}                                        \\ \hhline{ ~~----}
                                  & \multicolumn{1}{l|}{}                                  & \multicolumn{1}{l|}{Test Acc} & \multicolumn{1}{l|}{Challenge Acc} & \multicolumn{1}{l|}{Test Acc} & \multicolumn{1}{l}{Challenge Acc} \\ \hlineB{2}
\multirow{2}{*}{MobileNetV2}      & 50                                                     & 0.7365                        & 0.7151                             & 0.9836                        & 0.7402                            \\
                                  & 200                                                    & 0.7081                        & 0.6702                             & 0.9855                        & 0.6693                            \\ \hline
\multirow{2}{*}{EfficientNetB0}     & 50                                                     & 0.7035                        & 0.6812                             & 0.9873                        & 0.4331                            \\
                                  & 200                                                    & 0.7194                        & 0.7188                             & 0.9873                        & 0.4252                            \\ \hline
\multirow{2}{*}{VGG16}            & 50                                                     & 0.6881                        & 0.6432                             & 0.9909                        & 0.3701                            \\
                                  & 200                                                    & 0.7220                        & 0.6933                             & 0.9873                        & 0.3701                            \\ \hlineB{2}
\end{tabular}
\end{table*}

\begin{figure*} [h!]
	\centering
	\includegraphics[scale=0.5]{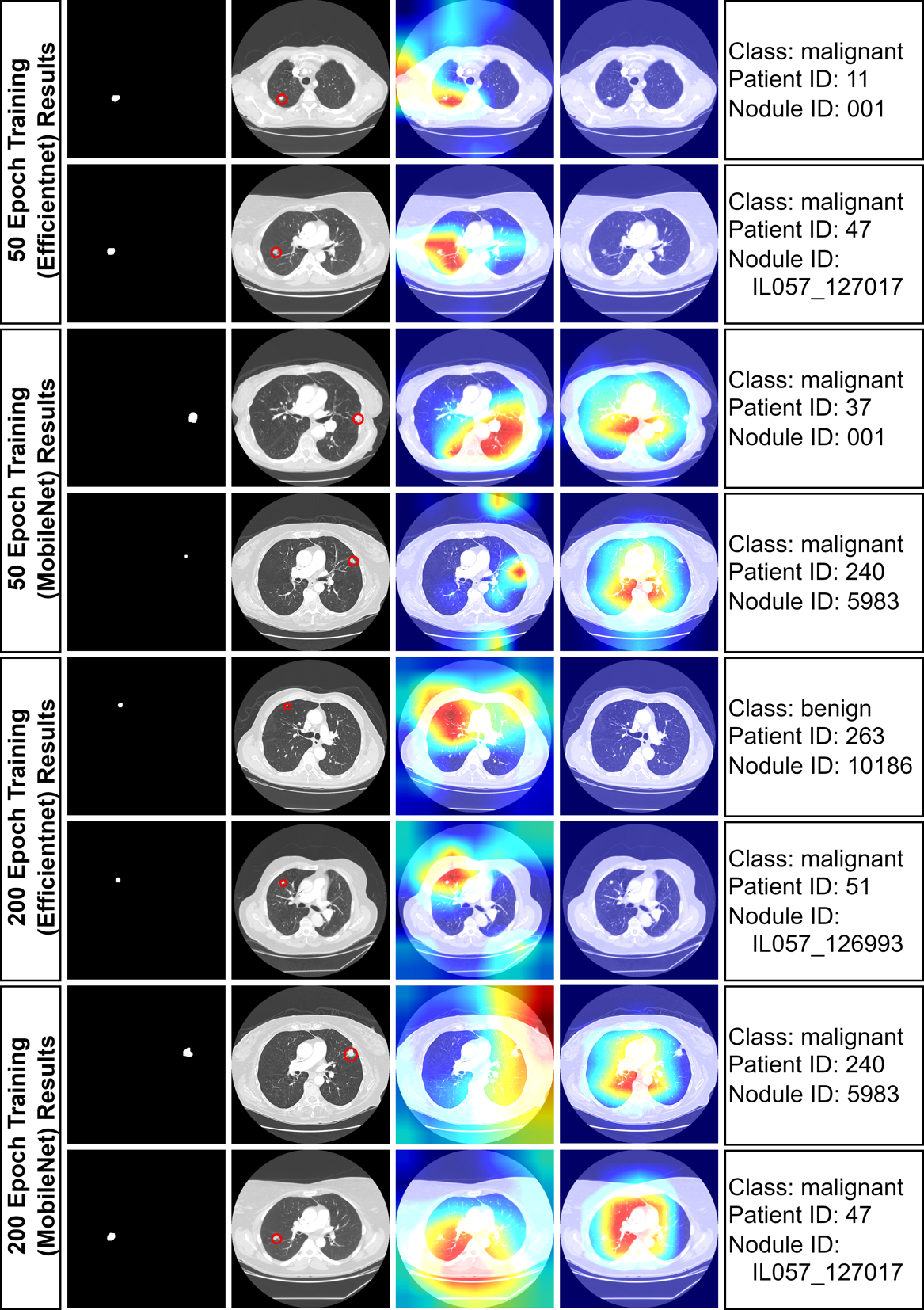}
	\caption{Heat-Map visualizations (CAMs) for eight randomly selected test images. The first column indicates masks of the nodules, the second column indicates original CT images, the third column indicates fair model CAMs, and the fourth column indicates unfair model CAMs. Nodules are pointed with the red circles in the original CT images.}
	\label{figsamples3}    
\end{figure*}

\section{Interpretability Analysis}

The use of heat maps, also referred to as Class Activation Maps (CAMs), is a common technique for visualizing the magnitude of a phenomenon through color-coded representations~\cite{Selvara2016, Yetgin2019}. In the context of deep neural networks (DNNs), which often operate as black boxes with limited interpretability, visualizing the decision-making process is crucial for assessing the fairness of the model. The creation of heat maps involves several steps, including preprocessing of the input image, prediction of the image class by the trained model, and calculation of gradients using both the output of the last convolutional layer and the output of the deep model. Neuron weights are then acquired via average pooling of these gradients in three axes. The values in each layer of the last convolutional block are subsequently multiplied by their corresponding neuron weights, and the average and maximum of these values are computed to generate the heat map. The heat map is then normalized, resized to the input image dimension, multiplied by 255, and subjected to color mapping before being combined with the original image. By highlighting the areas of an image that are most influential in the model's prediction, heat maps can provide insight into the internal workings of DNNs and their ability to perform complex tasks.

\begin{figure} [b!]
	\centering
	\includegraphics[scale=0.56]{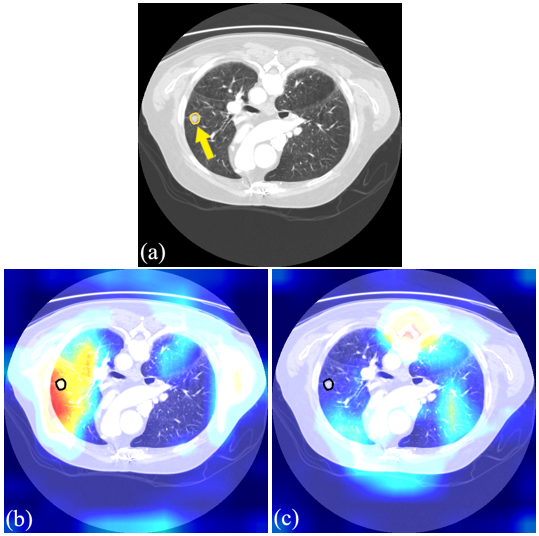}
	\caption{(a) Example of a CT image with a malignant nodule from LIDC/IDRI (with a yellow arrow showing nodule place); (b) corresponding fair model heat-map output; (c) corresponding unfair model heat-map output.} 
	\label{figsamplesthree}    
\end{figure}

In order to illustrate the interpretability analysis in detail, a set of 8 images was randomly selected and used to test both fair and unfair models. The aim was to use heat maps to identify the regions in the lungs that the models mainly use to make their final decisions. The results of these tests are presented in Figure~\ref{figsamples3}, with the red color indicating the strongest activations, using standard heat color maps from the OpenCV Python library. Upon reviewing the heat maps, it becomes clear that the unfair model produced malign predictions despite focusing on areas that are not even tumor regions. Conversely, the fair model was able to make malign predictions by focusing exclusively on the tumor regions. This finding demonstrates that the unfair models are not reliable, as they concentrate on areas that are not related to the tumor before making their final decisions. This lack of reliability would likely be amplified if the models were trained and tested on different patient images, as demonstrated by the fair model's test scores.

    Figure~\ref{figsamplesthree}-a shows a malignant CT image which is taken from a scan with an ID of 54 in the LIDC/IDRI dataset (tumor region is indicated). Once this image is tested in a fair and an unfair model, it is predicted correctly as malignant by both models. However, the reliability of these results becomes clear once the activation heat maps are overlaid on the CT image for the tested models. The heat map in Figure~\ref{figsamplesthree}-b shows that the regional activation (hence visual attention) of the fair model is high at and around the actual tumor region. On the other hand, Figure~\ref{figsamplesthree}-c, which shows the heat map from the unfair model, indicates no visual focus on the locations around the tumor region. It is argued that the model in Figure~\ref{figsamplesthree}-c gives a correct decision using an unreliable reasoning as a result of probable overfitting through unfair training.

\begin{figure*} [h]
	\centering
	\includegraphics[width=1\textwidth]{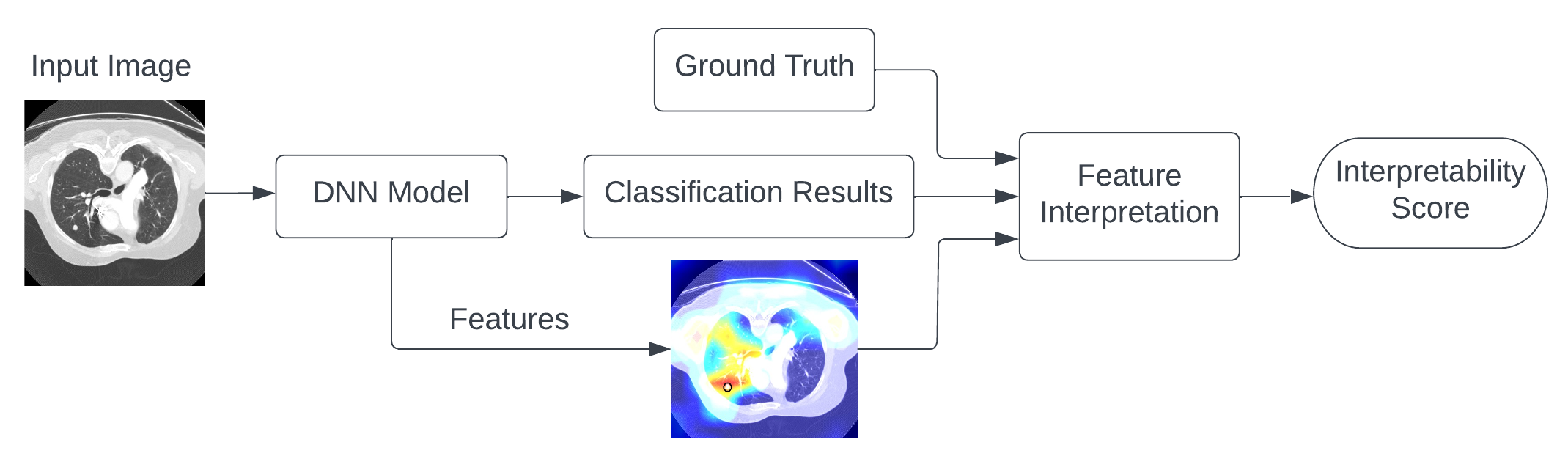}
	\caption{Flow chart of the interpretability analysis}
	\label{chart}    
\end{figure*}

To extend and generalize the findings from Figure~\ref{figsamplesthree}, the study employs two prominent interpretability score methodologies. Figure~\ref{chart} illustrates the general framework for conducting the interpretability analysis.

The first approach for the interpretability assessment focuses on attention heat map values that correspond to the tumor nodule region and compares them to the rest of the image. These values, which indicate higher activation and hence visual attention, can be either averaged inside the nodule regions, or the highest value inside the region can be considered as the attention value. Using the examples of the challenge images in Figure~\ref{figsamples3}, the mean and maximum heat map values inside the nodule regions using the unfair models are provided in Table~\ref{tab:my-table}. Clearly, both the maximum and the average heat maps inside the nodule regions using the unfair models are significantly lower than the values obtained using the fair models.

\begin{table*}[h!]
\caption{Heatmap values of several patients for Unfair and Fair cases with respect to nodule max and nodule mean criteria}
\label{tab:my-table}
\resizebox{0.7\textwidth}{!}{%
\begin{tabular}{|c|cc|cc|}
\hline
\multirow{2}{*}{Patient IDs} & \multicolumn{2}{c|}{Unfair}                                                   & \multicolumn{2}{c|}{Fair}                                          \\ \hhline{ ~----}
                             & \multicolumn{1}{c|}{Nodule Max} & Nodule Mean  & \multicolumn{1}{c|}{Nodule Max} & Nodule Mean \\ \hline
11                           & \multicolumn{1}{c|}{0}          & 0                & \multicolumn{1}{c|}{0.8882}    & 0.7908           \\ \hline
37                           & \multicolumn{1}{c|}{0.3824}  & 0.2997            & \multicolumn{1}{c|}{0.8030}    & 0.7076             \\ \hline
47a                           & \multicolumn{1}{c|}{0}       & 0                  & \multicolumn{1}{c|}{0.8677}    & 0.8116            \\ \hline
47b                           & \multicolumn{1}{c|}{0.6942} & 0.5901            & \multicolumn{1}{c|}{0.7444}    & 0.7046            \\ \hline
240a                          & \multicolumn{1}{c|}{0.3762} & 0.1990             & \multicolumn{1}{c|}{0.6319}    & 0.4167            \\ \hline
240b                          & \multicolumn{1}{c|}{0.6607} & 0.5152            & \multicolumn{1}{c|}{0.7529}    & 0.6972              \\ \hline
\end{tabular}%
}
\end{table*}
\begin{table*}[ht!]
\caption{Pearson and Spearman correlations of heatmap and ground truth nodule shapes for Unfair and Fair cases}
\label{tab:table4}
\resizebox{0.7\textwidth}{!}{%
\begin{tabular}{|c|cc|cc|}
\hline
\multirow{2}{*}{Patient IDs} & \multicolumn{2}{c|}{Unfair}                                                   & \multicolumn{2}{c|}{Fair}                                          \\ \hhline{ ~----}
                             & \multicolumn{1}{c|} {Pearson Corr} & Spearman Corr & \multicolumn{1}{c|}{Pearson Corr}  & Spearman Corr\\ \hline
11                           & \multicolumn{1}{c|} {0.0133}       &  0.0286       & \multicolumn{1}{c|}{0.0624}      & 0.0548        \\ \hline
37                           & \multicolumn{1}{c|} {0.0306}       &  0.0375       & \multicolumn{1}{c|}{0.0764}       & 0.0668        \\ \hline
47a                           & \multicolumn{1}{c|} {0.0112}       &  0.0183       & \multicolumn{1}{c|}{0.0563}       & 0.0561         \\ \hline
47b                           & \multicolumn{1}{c|} {0.0473}       &  0.0462       & \multicolumn{1}{c|}{0.0447}       & 0.0465           \\ \hline
240a                          & \multicolumn{1}{c|} {0.0053}       &  0.0132       & \multicolumn{1}{c|}{0.0327}       & 0.0353         \\ \hline
240b                          & \multicolumn{1}{c|} {0.0226}       &  0.0300       & \multicolumn{1}{c|}{0.0671}       & 0.0695           \\ \hline
\end{tabular}%
}
\end{table*}

The second approach for the interpretability assessment measures the structural similarity of the nodule regions and compares them to the structure of the shape obtained from the heat map image. It is argued that if these two shapes structurally match, it indicates a high interpretability score, and the DNN focuses on the nodule region with high attention. There are two well-known correlation techniques that measure the pixel layout similarity between two images; the Pearson and the Spearman correlation \cite{Pearson}. Pearson correlation evaluates the linear relationship between two images, whereas Spearman correlation is a more general measure that evaluates the monotonic relationship between two images. These classical correlation values are evaluated with the aim of finding the shape-wise relation between the focus heatmap values and binary morphological shape corresponding to the ground truth nodule label pixels. It is argued that a high correlation (closer to one) would indicate that the heatmap shows a correct focus to the nodule regions, whereas smaller correlation values would mean an incorrect, hence an uninterpretable focus. Table~\ref{tab:table4} shows Pearson and Spearman correlation values between nodule regions and heat map images for the set that was used in Figure~\ref{figsamples3}. Similar to the results in Table~\ref{tab:table4}, the stronger correlations between the nodule regions and the provided heat maps for the fair models indicate that the fair model causes a more reliable machine learning process as compared to the unfair model, where these correlation values are visibly lower.

\section{Discussions and Conclusion}

Lung cancer remains to be the leading cause of cancer-related deaths worldwide. Due to the importance of early detection of lung nodules and accurate differentiation between benign and malignant nodules in effective treatment and patient survival, the interest in fast and accurate application of computer-aided diagnosis is overwhelming. In recent years, ML and deep learning techniques have been widely used for the automatic classification of lung nodules in CT scans, providing a promising solution to improve the accuracy and efficiency of diagnosis over large volumes of data. The number of published articles regarding the use of ML or DNN-based methods for automatic nodule classification in CT images is well above several hundreds each year. Particularly, DNN-based approaches have shown great potential in the field of computer-aided diagnosis (CAD) for lung nodule classification from CT images. However, most of the deep learning-based classification techniques in the literature only focus on higher {\em reported} accuracy results, without considering the true reliability of the eventual system.

Our study has shown that patient-level separation is crucial in the training and testing of deep neural networks for lung nodule classification in CT images. Our findings indicate that careless image splitting without patient-wise separation in training and testing can lead to incorrect and unfair results that cannot be verified in new challenge datasets. On the other hand, patient-wise splitting in the training and testing process provides consistent, correct, and reliable results for accuracy percentages.

Moreover, the experimental results have also shown that patient-wise splitting in training and testing improves the interpretability of the constructed deep neural network by means of showing better attention to the activation values around the correct nodule regions. This improvement in interpretability was demonstrated using two different approaches: analysis of attention heat map values and correlation analysis between heat map images and the nodule regions.

Based on our findings, we recommend the following best practices for deep neural network training and testing for lung nodule classification in CT images:
\begin{itemize}
    \item Strictly separate the training, validation, and test datasets at the patient level to ensure reliable and interpretable results.
    \item Verify the interpretability of the trained networks by analyzing attention heat map values and correlation analysis between heat map images and the nodule regions.
    \item Report accuracy percentages for both overall performance and performance on new patient images to ensure the generalization of the deep neural network to new patients.
    \item Provide clear documentation of the dataset splitting methodology in any publications related to deep neural network training and testing for lung nodule classification in CT images.
\end{itemize}

The provided observations indicate that further care must be taken in ML and DNN applications of crucial medical applications such as benign/malignant classifications or diagnosis aid for achieving better reliability and real usability in medicine.


\end{document}